\documentclass{article}

\usepackage{microtype}
\usepackage{graphicx}
\usepackage{adjustbox}
\usepackage{subfigure}
\usepackage{booktabs} % for professional tables

\usepackage[accepted]{icml2024}
\usepackage{times}

\usepackage{amsmath}
\usepackage{amssymb}
\usepackage{mathtools}
\usepackage{amsthm}

\usepackage{comment}
\usepackage[hidelinks]{hyperref}

\date{}

\newcommand{\E}{\ensuremath{\mathbb{E}}}
\newcommand{\R}{\ensuremath{\mathbb{R}}}

\DeclareMathOperator{\Tr}{Tr}

\begin{document}

\twocolumn[
\icmltitle{Gated recurrent neural networks discover attention}
\icmlsetsymbol{equal}{*}
\icmlsetsymbol{senior}{$\dagger$}

\begin{icmlauthorlist}
\icmlauthor{Nicolas Zucchet}{equal,DINFK}
\icmlauthor{Seijin Kobayashi}{equal,DINFK}
\icmlauthor{Yassir Akram}{equal,DINFK}
\icmlauthor{Johannes von Oswald}{DINFK}
\icmlauthor{Maxime Larcher}{DINFK}
\icmlauthor{Angelika Steger}{senior,DINFK}
\icmlauthor{João Sacramento}{senior,DINFK}
\icmlaffiliation{DINFK}{Department of Computer Science, ETH Zürich}
\end{icmlauthorlist}

\icmlcorrespondingauthor{}{nzucchet, seijink, yakram, voswaldj, larcherm, asteger, rjoao@ethz.ch}
%\icmlcorrespondingauthor{Firstname2 Lastname2}{first2.last2@www.uk}
%\footnotesize{\texttt{\{nzucchet, seijink, yakram, voswaldj, larcherm, asteger, rjoao\}@ethz.ch}}

\icmlkeywords{Machine Learning, ICML}

\vskip 0.3in
]

\printAffiliationsAndNotice{\icmlEqualContribution}

\begin{abstract}
Recent architectural developments have enabled recurrent neural networks (RNNs) to reach and even surpass the performance of Transformers on certain sequence modeling tasks. These modern RNNs feature a prominent design pattern: linear recurrent layers interconnected by feedforward paths with multiplicative gating. Here, we show how RNNs equipped with these two design elements can exactly implement (linear) self-attention.
By reverse-engineering a set of trained RNNs, we find that gradient descent in practice discovers our construction. In particular, we examine RNNs trained to solve simple in-context learning tasks and find that gradient descent instills in our RNNs the same attention-based in-context learning algorithm. Our findings highlight the importance of multiplicative interactions in neural networks and suggest that certain RNNs might be unexpectedly implementing attention under the hood.
\end{abstract}

\section{Introduction}

Attention-based neural networks, most notably Transformers \citep{vaswani_attention_2017}, have rapidly become the state-of-the-art deep learning architecture, replacing traditional models such as multi-layer perceptrons, convolutional neural networks, and recurrent neural networks (RNNs). This is particularly true in the realm of sequence modeling, where once-dominating RNNs such as the long short-term memory \citep[LSTM;][]{hochreiter_long_1997} model and the related gated recurrent unit \citep[GRU;][]{cho_properties_2014} have been mostly replaced by Transformers.

Nevertheless, RNNs remain actively researched for various reasons, such as their value as models in neuroscience \citep{dayan_theoretical_2001}, or simply out of genuine interest in their rich properties as a dynamical system and unconventional computer \citep{jaeger_toward_2023}. Perhaps most importantly for applications, RNNs are able to perform inference for arbitrarily long sequences at a constant memory cost, unlike models based on conventional softmax-attention layers \citep{bahdanau_neural_2015}. This ongoing research has led to a wave of recent developments. On the one hand, new deep linear RNN architectures \citep{gu_efficiently_2022, orvieto_resurrecting_2023} have been shown to significantly outperform Transformers on challenging long-sequence tasks \citep[e.g.,][]{tay_long_2020} and on some language modelling tasks \cite{gu_mamba_2023}. On the other hand, many efficient linearized attention models have been developed, whose forward pass can be executed in an RNN-like fashion at a constant inference memory cost \citep[][]{tsai_transformer_2019,katharopoulos_transformers_2020,choromanski_rethinking_2021,schlag_linear_2021, fu_hungry_2023, sun_retentive_2023, yang_gated_2023}.

We present a unifying perspective on these two seemingly unrelated lines of work by providing a set of parameters under which gated RNNs become equivalent to any linearized self-attention, without requiring infinite number of neurons or invoking a universality argument. Crucially, our construction makes use of elementwise multiplications, which are ostensibly featured in different forms in recent deep linear RNN models. Turning to LSTMs and GRUs, which also include these multiplicative gating interactions, we find somewhat surprisingly that our results extend only to LSTMs. Moreover, the LSTM construction we provide requires a very specific configuration, which hints that the inductive bias towards attention-compatible configurations might be weaker for this architecture than for deep gated linear RNNs.

We then demonstrate that linear RNNs with multiplicative interactions, but not LSTMs and GRUs, can effectively implement our construction once trained, thus behaving as attention layers. Moreover, we find that such linear RNNs trained to solve linear regression tasks acquire an attention-based in-context learning algorithm. Incidentally, it has been shown that the very same algorithm is typically used by linear self-attention layers trained on this problem class \citep{von_oswald_transformers_2023,mahankali_one_2023,ahn_transformers_2023,zhang_trained_2023}. Our results thus challenge the standard view of RNNs and attention-based models as two mutually exclusive model classes and suggest that, through learning, RNNs with multiplicative interactions may end up encoding attention-based algorithms disguised in their weights.

\section{Background}

\subsection{Linear self-attention}
\label{subsec:background_lsa}

We study causally-masked linear self-attention layers that process input sequences $(x_t)_t$ with $x_t \in \mathbb{R}^d$ as follows:
\begin{equation}
    y_t = \left ( \sum_{t' \leq t} (W_V x_{t'}) (W_K x_{t'})^\top \right ) (W_Q x_t)
\end{equation}
In the previous equation, $W_V \in \mathbb{R}^{d\times d}$ is the value matrix, $W_K \in \mathbb{R}^{d \times d}$ the key matrix and $W_Q \in \mathbb{R}^{d\times d}$ the query matrix. We use square matrices throughout the paper for simplicity, but our findings extend to rectangular ones. As usually done, we call $v_t := W_V x_t$, $k_t := W_K x_t$ and $q_t := W_Q x_t$ the values, keys and queries. The output vector $y_t$ has the same dimension as the input, that is $d$. Such linear self-attention layers can be understood as a linearized version of the softmax attention mechanism \citep{bahdanau_neural_2015} in use within Transformers \citep{vaswani_attention_2017}. Yet, they operate in a very different regime than softmax layers, which have unbounded memory. Attention layers commonly combine different attention heads; we focus on a single one here for simplicity.

In a linear self-attention layer, information about the past is stored in an effective weight matrix $W^\mathrm{ff}_t := \sum_{t'} v_{t'} k_{t'}^\top$ that will later be used to process the current query $q_t$ through $y_t = W^\mathrm{ff}_t q_t$. At every timestep, $W_t^\mathrm{ff}$ is updated through the rule $W_t^\mathrm{ff} = W_{t-1}^\mathrm{ff} + v_t k_t^\top$, which is reminiscent of Hebbian learning \citep{schmidhuber_learning_1992, schlag_linear_2021} and leads to faster inference time \citep{katharopoulos_transformers_2020, choromanski_rethinking_2021, shen_efficient_2021, peng_random_2021} than softmax self-attention.

\subsection{Gated recurrent neural networks}
\label{sec:background_gated_RNN}

In this paper, we focus our analysis on a simplified class of gated diagonal linear recurrent neural networks. They implement bilinear input $g^\mathrm{in}$ and output gating $g^\mathrm{out}$ that multiplies a linear transformation $W_\mathrm{x}^\mathrm{in/out} x_t$ of the input with a linear gate $W_\mathrm{m}^\mathrm{in/out} x_t$: $g^\mathrm{in/out}(x_t) = (W_\mathrm{m}^\mathrm{in/out} x_t) \odot (W_\mathrm{x}^\mathrm{in/out} x_t)$. Here, $\odot$ is the elementwise product. The class of gated networks we consider satisfies
\begin{equation}
    \label{eq:gated_rnn}
    h_{t+1} = \lambda \odot h_t + g^\mathrm{in} (x_t), ~~~ y_t = D g^\mathrm{out}(h_t).
\end{equation}
In the previous equation, $\lambda$ is a real vector, $x_t$ is the input to the recurrent layer, $h_t$ the hidden state, and $D$ a linear readout. This simplified class makes connecting to attention easier while employing similar computational mechanisms as standard gated RNNs architectures.

This class is tightly linked to recent deep linear RNN architectures and shares most of its computational mechanisms with them. While linear diagonal recurrence might be seen as a very strong inductive bias, many of the recent powerful deep linear RNN models adopt a similar bias \citep{gupta_diagonal_2022, smith_simplified_2023, gu_mamba_2023}, and it has been shown to facilitate gradient-based learning \citep{orvieto_resurrecting_2023, zucchet_online_2023}. Those architectures often use complex-valued hidden states in the recurrence; we only use its real part here. Some of those works employ a GLU \citep{dauphin_language_2017} after each recurrent layer, with $\mathrm{GLU}(x) = \sigma(W_\mathrm{m} x_t) \odot W_\mathrm{x} x_t$ with $\sigma$ the sigmoid function. The gating mechanism we consider can thus be interpreted as a linearized GLU. We can recover \eqref{eq:gated_rnn} by stacking two layers: the GLU in the first layer acts as our input gating, and the one in the second as output gating. Alternatively, architectures like Mamba \cite{gu_mamba_2023} uses input-dependent matrices as projection to the hidden state instead of the input gating. Multiplying such matrices with the input itself thus results in a multiplicative gating. Its output gating mechanism is slightly different as one of the branch takes the input of the recurrent layer as input, instead of the hidden state. We include a more detailed comparison in Appendix~\ref{app:comparison_architectures}. In the rest of the paper, we will use the LRU layer \citep{orvieto_resurrecting_2023} as the representative of the deep linear RNN architectures because of its simplicity.

LSTMs can operate in the regime of Equation \ref{eq:gated_rnn}, but this requires more adaptation. First, the recurrent processing is nonlinear and involves more steps than are captured in \eqref{eq:gated_rnn}. Second, gating occurs in different parts of the computation and depends on additional variables. We compare in more details this architecture and the one of Equation \ref{eq:gated_rnn} in Appendix~\ref{app:comparison_architectures}, showing that LSTMs can implement \eqref{eq:gated_rnn} when stacking two layers on top of each other. We additionally show that GRUs cannot do so.

\begin{figure*}
    \centering    \includegraphics{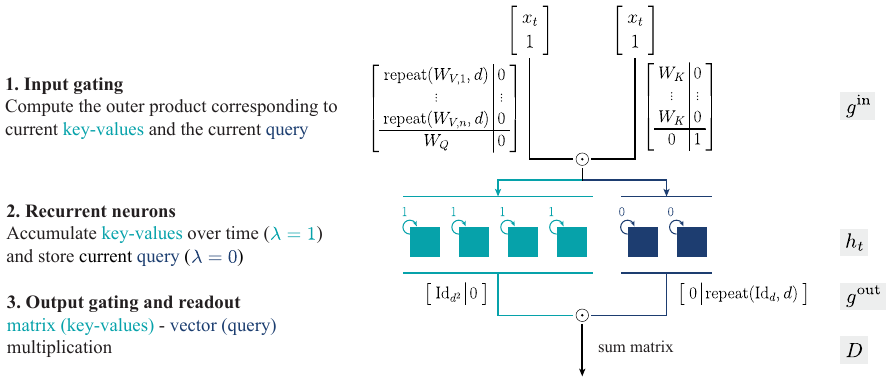}
    \caption{An example of a diagonal linear gated recurrent neural network that implements the same function as a linear self-attention layer with parameters $(W_V, W_K, W_Q)$ and input dimension $d$, as described in Section~\ref{sec:construction}. Inputs are processed from top to the bottom. We do not use biases so we append 1 to the input vector $x_t$ to be able to send queries to the recurrent neurons. We use $\mathrm{repeat}(A, n)$ to denote that the matrix $A$ is repeated $n$ times on the row axis and $W_{V,i}$ is the $i$-th row of the $W_V$ matrix. The bars within the matrices separate the different kinds of inputs/outputs. Digits in matrices denote column vectors appropriately sized. The readout matrix $D$ appropriately sums the elementwise products between key-values and queries computed after the output gating $g^\mathrm{out}$. Exact matrix values can be found in Appendix~\ref{app:explicit-construction}.}
    \label{fig:construction}
\end{figure*}

\section{Theoretical construction}
\label{sec:construction}

As highlighted in the previous section, our class of gated RNNs and linear self-attention have different ways of storing past information and using it to modify the feedforward processing of the current input. The previous state $h_t$ acts through a bias term $\lambda \odot h_t$ that is added to the current input $g^\mathrm{in}(x_t)$ in gated RNNs, whereas the linear self-attention recurrent state $W^\mathrm{ff}_t$ modifies the weights of the feedforward pathway. We reconcile these two mismatched views of neural computation in the following by showing that gated RNNs can implement linear self-attention.

In this section, we demonstrate how a gated recurrent layer followed by a linear readout as in Equation~\ref{eq:gated_rnn} can implement any linear self-attention layer through a constructive proof. In particular, our construction only requires a finite number of neurons to exactly match the desired function, therefore providing a much stronger equivalence result than more general universality of linear recurrent networks theorems \citep{boyd_fading_1985, grigoryeva_universal_2018, orvieto_universality_2023}, which hold in the limit of infinitely many recurrent neurons.

\subsection{Key ideas} 

Our construction comprises three main components: First, the input gating $g^\mathrm{in}$ is responsible for generating the elementwise products between the keys and values, as well as the queries. Then, recurrent units associated with key-values accumulate their inputs with $\lambda = 1$, whereas those receiving queries as inputs return the current value of the query, hence $\lambda = 0$. Lastly, the output gating $g^\mathrm{out}$ and the final readout layer $D$ are in charge of multiplying the flattened key-value matrix with the query vector. We illustrate our construction and provide a set of weights for which the functional equivalence holds in Figure~\ref{fig:construction}. Crucially, the key-values in a linear self-attention layer are the sum of degree two polynomials of each previous input. Input gating mechanism and perfect memory units ($\lambda = 1$) are needed to replicate this behavior within a gated recurrent layer. Similarly, output gating is required to multiply key-values with the queries.

\subsection{On the number of neurons needed}
\label{subsec:number_neurons_construction}

The construction of Figure~\ref{fig:construction} requires $d^2 + d$ hidden neurons to store all the entries of the $d\times d$ key-value matrix and of the query vector of size $d$. While this construction is arguably the most intuitive, it is not optimal in terms of number of neurons used. Knowing the exact minimal number of neurons is fundamental for understanding which solution the network learns. Therefore, we detail how we can make our construction more compact in the following. We leverage two insights: First, any combination of key and query matrices for which $(W_K^\top W_Q)$ is fixed leads to the same function in the linear self-attention layer. We can thus assume that the key and value matrices are equal, as taking the key matrix to be equal to $W_V$ and changing the query matrix to be $W_V^{-\top} W_K^\top W_Q$ does not change the behavior of the attention layer. Second, when the key and value matrices are equal, the key-value matrix is symmetric and, therefore, only requires $d(d+1)/2$ elements to be represented. This implies that, when the value matrix is invertible, the minimal number of hidden neurons our gated RNN needs to store key-values is in fact $d(d+1)/2 + d$. In Section~\ref{sec:gatedRNN_identification}, we show that learned RNNs find this solution.

Alternatively, it is also possible to reduce the construction size when the weight matrices of the teacher attention layer are of low rank. In this case, we still have a quadratic scaling of the required numbers of recurrent neurons, but this time in the rank of the different matrices instead of the entire dimension. The detailed derivation can be found in Appendix~\ref{app:low-rank-teacher}.

Overall, the output gating requires $\mathcal{O}(d^2)$ input and output entries for the gated RNN to match a linear self-attention layer. The RNN thus requires $\mathcal{O}(d^4)$ parameters in total, with a lot of redundancy, significantly more than the $3d^2$ parameters of the linear self-attention layer. We note that changing the output gating to a side one is possible, c.f. Appendix~\ref{app:side_gating_construction}, reducing the number of required parameters to $\mathcal{O}(d^3)$.

Given the high parameter redundancy, it comes as no surprise that numerous equivalent configurations exist within the gated RNN we study. For instance, linear gating is invariant under permutations of rows between its two matrices and under multiplication-division of these two rows by a constant. Left-multiplying $W_Q$ in the input gating by any invertible matrix $P$, and subsequently reading out the hidden neurons with $\lambda = 0$ through $\mathrm{repeat}(P^{-1}, d)$, also does not alter the network's output. Several other invariances exist, making exact weight retrieval nearly impossible. These considerations will be of practical use when we will reverse engineer the function encoded by trained recurrent networks in Section~\ref{subsec:teacher-identification}.

\subsection{Implications for existing classes of RNNs}

We conclude this section by commenting on whether similar insights hold for more realistic gated RNNs architectures.

The LRU architecture is close to \eqref{eq:gated_rnn} but only contains output gating through a GLU layer. Stacking two LRU layers on top of each other enables the output gating of the first layer to act as the input gating for the second layer and, therefore, implement the mechanism we highlighted in the previous sections to mimick attention. Intuitively, adding an input GLU would bias the LRU towards linear self-attention as one layer would now enough to implement it. We will later confirm that this indeed improves the LRU ability to mimick linear self-attention, as well as boost its performance on certain tasks. The Mamba block has a stronger inductive bias towards attention due to the presence of a side gating querying the memory stored in the recurrent state. Interestingly, it has been found that removing the input dependence of the matrix projecting to the hidden state is detrimental to performance \cite{gu_mamba_2023}. This decreases the inductive bias towards linear self-attention, which might partly explain the performance drop.

As noted in Section~\ref{sec:background_gated_RNN}, LSTMs and GRUs are further away from our simplified gated RNN model. However, one single LSTM layer can implement linear self-attention, but stacked GRU layers cannot. Let us briefly summarize the argument behind these results. The LSTM layer has a sophisticated input gating mechanism that gates a candidate cell state based on the current input and previous state. The gate and the candidate cell state depend, among other things, on the current input. This mechanism can thus play a similar role to $g^\mathrm{in}$ and implement the key-value outer product. The recurrence of the cell state can be set to perfectly integrate key-values, by setting the forgetting gate accordingly. Finally, the output gate modulates the current cell state, which contains the accumulated key-values. Setting the output gate to encode the query enables computing the desired result. We note that the output gating differs from $g^\mathrm{out}$: it multiplies transformations of the cell state and the input instead of the input only. This property makes it possible to implement attention within one layers, where as two layers are required for our gated RNN model~\eqref{eq:gated_rnn}. While the GRU layer takes many of the computational elements from the LSTM, it cannot implement attention as it has no mechanism to compute multiply keys and values.

We refer the reader to Appendix~\ref{app:comparison_architectures} for more details.

\begin{figure*}[ht]
    \centering
    \includegraphics{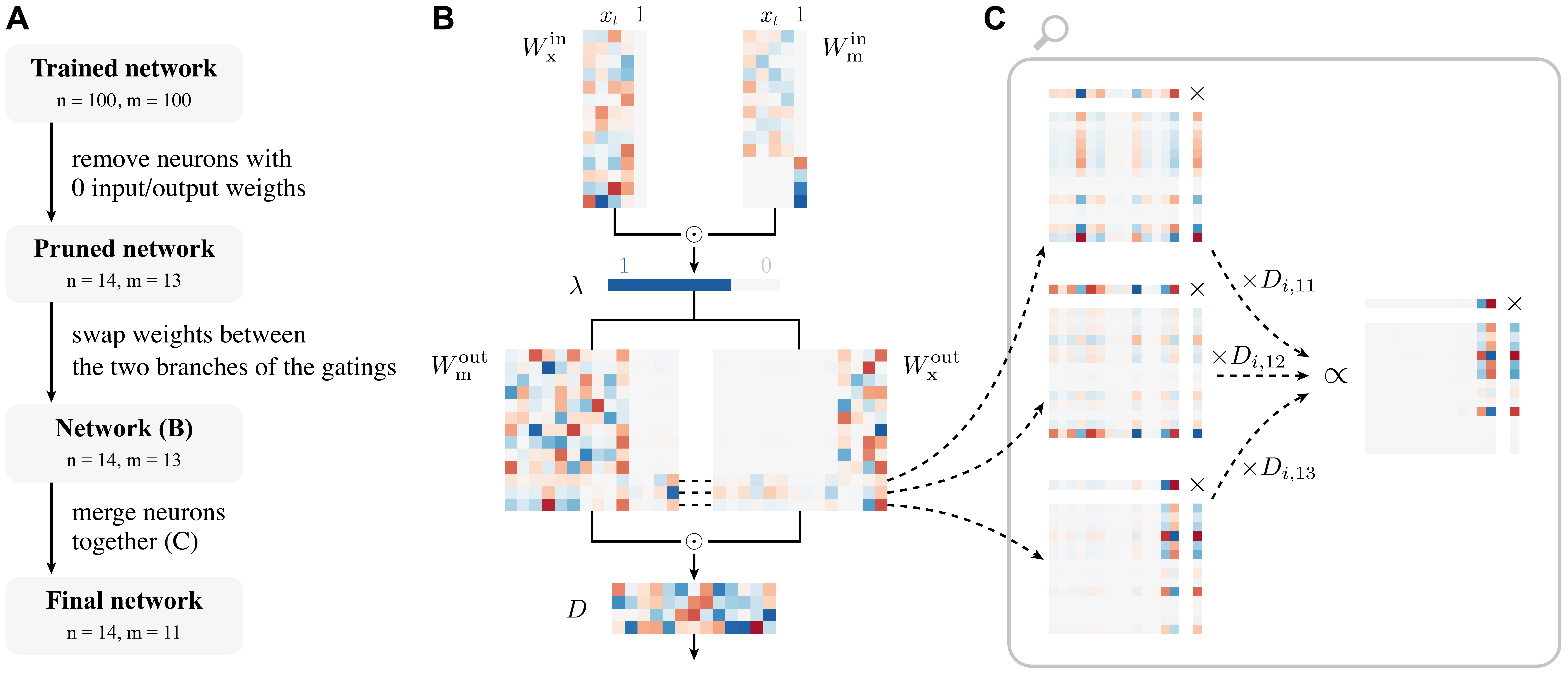}
    \caption{In our teacher-student experiment of Section~\ref{subsec:teacher-identification}  ($d=4$), the structure of the weights of the RNN after learning matches the one of our compact construction, c.f. Section~\ref{sec:construction}. \textbf{(A)} Summary of the post-processing we apply to the trained network weights. The number of recurrent neurons is denoted $n$, and the number of neurons after the output gating is denoted $m$. \textbf{(B)} Only recurrent neurons with perfect memory ($\lambda=1$, dark blue) or no memory at all ($\lambda=0$, light grey) influence the output, consistently with the theory. The block structure of the different weight matrices almost perfectly match the one of our construction, c.f. Figure~\ref{fig:construction} \textbf{(C)} The last three output neurons of the output gating are functionally equivalent to a single neuron whose input weights match the structure of the rest of the output gating weights. This can be achieved by representing each such neuron as an outer product (left part) which will later be combined by the readout matrix $D$. The combined kernels are rank 1 and proportional to each other. They can thus be expressed as the same outer product (right part). In all the matrices displayed here, zero entries are shown in light grey, blue denotes positive entries, and red negative ones.}
    \label{fig:result-learning}
\end{figure*}

\section{Gated RNNs learn to mimic attention}
\label{sec:gatedRNN_identification}

We now demonstrate that gated RNNs learn to implement linear self-attention and comprehend how they do so. In this section, a student RNN is tasked to reproduce the output of a linear self-attention layer. Appendix~\ref{app:teacher_student} contains detailed descriptions of all experiments performed in this section. Importantly, each sequence is only presented once to the network.

\subsection{Teacher identification}
\label{subsec:teacher-identification}

In our first experiment, we train a student RNN ($|x| = 4$, $|h|=100$ and $|y|=4$) to emulate the behavior of a linear self-attention layer with weights sampled from a normal distribution and inputs $x_t$ sampled i.i.d. from a normal distribution. The low training loss, reported in Table~\ref{tab:identificability}, highlights that the student's in-distribution behavior aligns with the teacher's. However, this is insufficient to establish that the student implements the same function as the teacher. The strategy we adopt to show functional equivalence is as follows: First, we observe that only perfect memory neurons ($\lambda = 1$) and perfect forget neurons ($\lambda=0$) influence the network output. Additionally, each of these groups of neurons receives all the information needed to linearly reconstruct resp.~the key-values and the queries from the input (Table~\ref{tab:identificability} Score KV and Score Q columns). Finally, we show that the output gating and the decoder matrix accurately multiply accumulated key-values with current queries, leading to proper identification of the teacher self-attention function, even outside the training distribution (Table~\ref{tab:identificability} Polynomial distance).

\begin{figure*}
    \centering
    \includegraphics{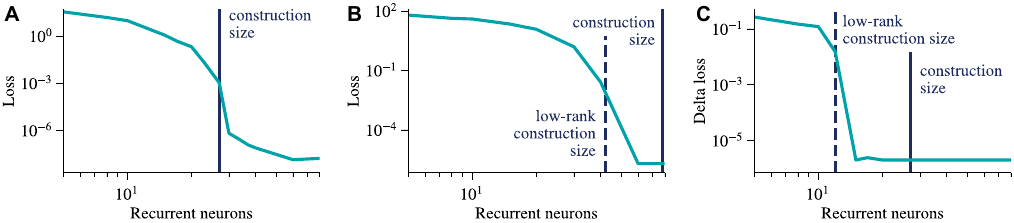}
    \caption{Gated RNNs learn compressed representations when possible. In the teacher-student experiment of Section~\ref{sec:gatedRNN_identification} \textbf{(A, B)}, the gated RNN identifies the teacher function under mild overparametrization. When the attention layer weights are low rank \textbf{(B)} the RNN learns a more compressed representation than what it would do when they are full rank \textbf{(A)}. \textbf{(C)} In the linear regression task of Section~\ref{sec:ICL}, the gated RNN behaves similarly to the optimal linear attention layer for that task, as the difference between their losses (delta loss) goes to 0. Moreover, the RNN discovers the same low-rank structure as this attention layer.}
    \label{fig:varying_student}
\end{figure*}

After the learning process, a significant part of the weights in the input and output gating and the readout becomes zeros. We can thus prune neurons with input or output weights that are entirely zeros, thereby preserving the network's function. By doing so, we can remove $86$ out of the $100$ hidden neurons and $87$ out of the $100$ pre-readout neurons. After having permuted rows in the two gating mechanisms and reordered hidden neurons, we plot the resulting weights on Figure~\ref{fig:result-learning}.B. Consistently with our construction, only recurrent neurons with $\lambda = 0$ or $\lambda = 1$ contribute to the network's output. The key-values neurons receive a polynomial of degree $2$, as $g^\mathrm{in}$ is a bilinear form, without any term of degree $1$ as the last column of $W_\mathrm{m}^\mathrm{in}$ and $W_\mathrm{x}^\mathrm{in}$ is equal to zero for those units. Similarly, the query neurons receive a polynomial of degree $1$. The learning process discovers that it can only use $d(d+1)/2 = 10$ neurons to store key-values, similar to our optimal construction. We show in Table~\ref{tab:identificability} that it is possible to linearly reconstruct the key-values from those $10$ neurons perfectly, as well as the queries from the $4$ query neurons. By combining this information with the fact that the $\lambda$s are zeros and ones, we deduce that the cumulative key-values $\sum_{t'\leq t} v_{t'} k_{t'}^\top$ can be obtained linearly from the key-values' hidden neurons, and the instantaneous queries $q_t$ from the query neurons.

\begin{table}[]
    \centering
    \begin{adjustbox}{width=\columnwidth,center}
    \begin{tabular}{cccccc}
        \toprule
         Loss & Score KV & Score Q & Polynomial distance\\
         \midrule
        $4.97 \times 10^{-8}$ & $4.52 \times 10^{-8}$ & $2.06 \times 10^{-10}$ & $3.73 \times 10^{-4}$ \\
        \bottomrule
    \end{tabular}
    \end{adjustbox}
    \caption{Gated RNNs implement the same function as a linear self-attention layer in our teacher-student experiment (Section~\ref{subsec:teacher-identification}). The KV and Q scores are equal to one minus the $R^2$ score of the linear regression that predicts key-values and queries from resp. the perfect memory neurons (those whose $\lambda=1$) and perfect forget neurons ($\lambda = 0$). The polynomial distance is the L2 distance between the coefficients of the degree-4 polynomial that describes the instantaneous processing of the (optimal) linear self-attention layer and the trained RNN.
    }
    \label{tab:identificability}
\end{table}

Additionally, the output gating combined with the linear readout can multiply the key-values with the queries. Since we have already confirmed that the temporal processing correctly accumulates key-values, our focus shifts to proving that the instantaneous processing of the gated RNN matches the one of the attention layer across the entire input domain. Given that both architectures solely employ linear combinations and multiplications, their instantaneous processing can be expressed as a polynomial of their input. The one of linear self-attention, $(W_Vx)(W_Kx)^\top(W_Qx)$, corresponds to a polynomial of degree $3$, whereas the one of the gated RNN, $g^\mathrm{out}(g^\mathrm{in}(x))$, corresponds to one of degree $4$. By comparing these two polynomials, we can compare their functions beyond the training domain. For every one of the four network outputs, we compute the coefficients of terms of degree $4$ or lower of their respective polynomials and store this information into a vector. We then calculate the normalized Euclidean distance between these coefficient vectors of the linear self-attention layer and the gated RNN, and report the average over all 4 output units in Table~\ref{tab:identificability}. The evidence presented so far enables us to conclude that the student network has correctly identified the function of the teacher.

While the majority of the weights depicted in Figure~\ref{fig:result-learning}.A conform to the block structure characteristic of our construction, the final three rows within the output gating matrices deviate from this trend. As shown in Figure~\ref{fig:result-learning}.B, these three rows can be combined into a single row matching the desired structure. More details about this manipulation can be found in Appendix~\ref{app:compression_output_gating}.

\subsection{Identification requires mild overparametrization}

The previous experiment shows that only a few neurons in a network of $100$ hidden neurons are needed to replicate the behavior of a self-attention layer whose input size is $d$. We therefore wonder if identification remains possible when decreasing the number of hidden and pre-output gating neurons the student has. We observe that mild overparametrization, around twice as many neurons as the actual number of neurons required, is needed to reach identification. We report the results in Figure~\ref{fig:varying_student}.A.

\subsection{Nonlinearity makes identification harder}
\label{subsec:TS-comparison}

We now move away from our simplified class of gated RNNs and seek to understand how our findings apply to LSTMs, GRUs, and LRUs. We use the following architecture for those three layers: a linear embedding layer projects the input to a latent representation, we then repeat the recurrent layer once or twice, and finally apply a linear readout. While those layers are often combined with layer normalization, dropout, or skip connections in modern deep learning experiments, we do not include any of those here to stay as close as possible to the teacher's specifications. In an LRU layer, the input/output dimension differs from the number of different neurons; we here set all those dimensions to the same value for a fair comparison with LSTMs and GRUs. We compare these methods to the performance of our simplified gated RNNs, with both diagonal (as in Equation~\ref{eq:gated_rnn}) and dense linear recurrent connectivity.

\begin{figure*}
    \centering
    \includegraphics{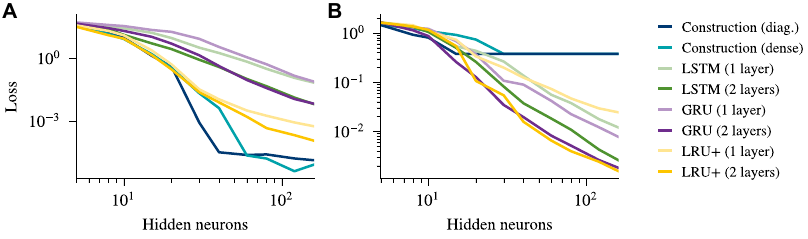}
    \caption{Comparison of the test loss obtained by different gated recurrent networks architectures in \textbf{(A)} the teacher-student task of Section~\ref{sec:gatedRNN_identification} and \textbf{(B)} the in-context linear regression task of Section~\ref{sec:ICL}. The construction baseline corresponds to the gated RNN of Eq.~\ref{eq:gated_rnn}, with diagonal or dense connectivity. We use the default implementation of LSTMs and GRUs, and slightly modify the LRU architecture to reflect our construction better. Non-linearity improves the in-context learning performance but deteriorates the ability to mimic attention.}
    \label{fig:comparison}
\end{figure*}

We report the results in Figure~\ref{fig:comparison}.A for inputs of dimension $d=6$. While diagonal connectivity provides a useful inductive bias to learn how to mimic linear self-attention, it is not absolutely needed as changing the recurrence connectivity to be dense does not significantly affect performance. It is theoretically possible to identify the teacher with one LSTM layer. However, gradient descent does not find such a solution and the performance of LSTMs is close to that of GRUs that cannot implement attention. Motivated by the construction of Section~\ref{sec:construction}, we slightly modify the LRU architecture (LRU+) and add a nonlinear input gating to the already existing output gating. We find that this modification significantly improves the ability of a LRU layer to mimic attention. Appendix~\ref{app:teacher_student} contains experiments that extensively compare different LRU architectures, as well as comparisons that take into account the number of parameters of the different architectures. Additionally, we provide results confirming that multiplicative interactions are fundamental for mimicking attention: replacing gating with a 1-hidden layer MLP with the same number of parameters significantly deteriorates performance. 

\section{Attention-based in-context learning emerges in trained RNNs}
\label{sec:ICL}

The previous section shows that gated RNNs learn to replicate a given linear self-attention teacher. We now demonstrate that they can find the same solution as linear self-attention when both are learned. To that end, we study an in-context regression task in which the network is shown a few input-output pairs and later has to predict the output value corresponding to an unseen input. Linear self-attention is a particularly beneficial inductive bias for solving this task. When the input-output mapping is linear, \cite{von_oswald_transformers_2023} have shown that linear self-attention implement one step of gradient descent.

\subsection{In-context linear regression}

Linear regression consists in estimating the parameters $W^*\in R^{d_y\times d_x}$ of a linear model $y = W^* x$ from a set of observations $\{(x_t, y_t)\}_{t=1}^T$ that satisfy $y_t = W^*x_t$. The objective consists in finding a parameter $\hat{W}$ which minimizes the squared error loss $L(W) = \frac{1}{2T}\sum_{t=1}^T \|y_t - Wx_t \|^2$. Given an initial estimate of the parameter $W_0$, one step of gradient descent on $L$ with learning rate $T\eta$ yields the weight change
\begin{equation}
\Delta W_0 = \eta\sum_{t=1}^T (y_t - W_0 x_t)x_t^\top.
\end{equation}
In the in-context version of the task, the observations $(x_t, y_t)_{1\leq t \leq T}$ are provided one after the other to the network, and later, at time $T+1$, the network is queried with $(x_{T+1}, 0)$ and  its output regressed against $y_{T+1}$. Under this setting, \citet{von_oswald_transformers_2023} showed that if all bias terms are zero, a linear self-attention layer learns to implement one step of gradient descent starting from $W_0 = 0$ and predict through
\begin{equation} \label{eq:gd_solution}
\hat{y}_{T+1} = (W_0 + \Delta W_0)x_{T+1} = \eta\sum_{t=1}^T y_t x_t^\top x_{T+1}.
\end{equation}
In the following, we show that gated RNNs also learn to implement the same algorithm and leverage the sparse structure of the different attention matrices corresponding to gradient descent to learn a more compressed representation than the construction one.

\subsection{Gated RNNs learn to implement gradient descent}

We now train gated RNNs as in Equation~\ref{eq:gated_rnn} to solve the in-context linear regression task, see Appendix~\ref{app:ICL-experimental-details} for more details. We set the number of observations to $T=12$ and set the input and output dimensions to $3$ so that $d=6$. Once learned, the RNN implements one step of gradient descent with optimal learning rate, which is also the optimal solution one layer of linear self-attention can find~\citep{mahankali_one_2023}. Several pieces of evidence back up this claim: the training loss of RNN after training ($0.0945$) is almost equal to the one of an optimal step of gradient descent ($0.0947$) and the trained RNN implements the same instantaneous function, as the polynomial analysis of Table~\ref{tab:gd_polynomial} reveals.

\begin{table}[]
    \centering
    \begin{tabular}{cccc}
        \toprule
          Term & RNN & GD\\
         \midrule
          $x_1^2y_1$ & $6.81 \times 10^{-2} \pm 8.52 \times 10^{-5}$ & $6.76 \times 10^{-2}$ \\
          $x_2^2y_1$ & $6.82 \times 10^{-2} \pm 6.40 \times 10^{-5}$ & $6.76 \times 10^{-2}$ \\
          $x_3^2y_1$ & $6.82 \times 10^{-2} \pm 5.56 \times 10^{-5}$ & $6.76 \times 10^{-2}$ \\
          residual   & $1.35 \times 10^{-3} \pm 1.97 \times 10^{-4}$ & 0 \\
        \bottomrule
    \end{tabular}
    \caption{Gated RNNs implement gradient descent in the in-context linear regression task of Section~\ref{sec:ICL}. Here, the input (resp. output) at time $t$ is denoted as $x_t=(x_{t,1},x_{t,2},x_{t,3})^\top$ (resp. $y_t=(y_{t,1},y_{t,2},y_{t,3})$). The instantaneous function for each output neuron can implement a polynomial of degree 4 in these terms. The table shows the coefficients of the polynomial implemented by the first output neuron of a trained RNN on the in-context linear regression task. Interestingly, the only terms without negligible coefficients (averaged over 4 seeds) are $(x_1)^2y_1, (x_3)^2y_1, (x_3)^2y_1$. The polynomial is virtually identical to that of one optimal step of gradient descent. The optimal GD learning rate is obtained analytically ($\eta^* = (T+d_x-1/5)^{-1}$), c.f. Appendix~\ref{sec:optimal_lr}. The residual norm measures the norm of the polynomial coefficients, excluding the ones appearing in the table. }
    \label{tab:gd_polynomial}
\end{table}

Linear self-attention weights implementing gradient descent have a very specific low-rank structure~\citep{von_oswald_transformers_2023}. To test whether the network learned our corresponding compressed construction, we vary the gated RNN size and report in Figure~\ref{fig:varying_student}.C the difference between the final training loss and the loss obtained after one optimal gradient descent step. We observe a similar transition from high to low low than in the teacher-student experiment, this time happening around the number of recurrent neurons prescribed by our low-rank construction. Gated RNNs thus learn a more compressed representation than the one naively mimicking self-attention. This result provides some hope regarding the poor $\mathcal{O}(d^4)$ scaling underlying our construction: in situations that require an attention mechanism with low-rank $(W_V, W_K, W_Q)$ matrices, gated RNNs can implement attention with far fewer neurons. A precise understanding of how much compression is possible in practical scenarios requires further investigation.

In Appendix~\ref{app:assoc-recall}, we provide an additional set of results focusing on associative recall, an in-context task where the goal is to memorize (and then retrieve) associations between pairs of inputs presented in sequence \citep{fu_hungry_2023}. This may be viewed as a simple instance of in-context classification, which does not require  generalization. As for linear regression, we find that trained gated RNNs discover an algorithm similar to the one employed by linear self-attention.

\subsection{Nonlinear gated RNNs are better in-context learners than one step gradient descent}

Finally, as a side question, we compare the ability to learn in context of the nonlinear gated RNN architectures that are LSTMs, GRUs and LRUs. Although not the main focus of our paper, this allows us to put our previous results in perspective. In particular, we are interested in understanding if similarity with attention correlates with in-context learning performance, as attention has been hypothesized to be a key mechanism for in-context learning \citep{olsson_-context_2022, garg_what_2022, von_oswald_transformers_2023}. We report our comparison results in Figure~\ref{fig:comparison}.B, measuring the loss on weights $W^*$ drawn from a distribution with double the variance of the one used to train the model.

Overall, we find that nonlinearity greatly helps and enables nonlinear gated RNN architectures to outperform one gradient descent step when given enough parameters, suggesting that they implement a more sophisticated mechanism.
Surprisingly, while the GRU is the architecture that is the furthest away from attention, it performs the best in the task. Within the different LRU layers we compare, we find a high correlation between in-context learning abilities and closeness to attention, c.f. Figure~\ref{fig:extensive_comparison} in the Appendix. In particular, we observe a massive performance improvement from the vanilla LRU architecture to the ones additionally including input gating to match our construction more closely. Once again, replacing the GLU by a MLP leads to a great decrease in performance.

\section{Discussion}

Our study reveals a closer conceptual relationship between RNNs and attention-based architectures than commonly assumed. We demonstrate that gated RNNs can theoretically and practically implement linear self-attention, bridging the gap between these two architectures. Moreover, while Transformers have been shown to be powerful in-context learners \citep{brown_language_2020, chan_data_2022}, we find that RNNs  excel in toy in-context learning tasks and that this performance is partly uncorrelated with the architecture inductive bias toward attention. This highlights the need for further investigations on the differences between RNNs and Transformers in controlled settings, as also advocated by \cite{garg_what_2022}.

Our results partly serve as a negative result: implementation of attention is possible but requires squaring the number of parameters attention has. We have shown that gated RNNs can leverage possible compression, but understanding whether real-world attention mechanisms lie in this regime remains an open question. Yet, our work is of current practical relevance as it provides a framework that can guide future algorithmic developments, as we exemplify in Appendix~\ref{app:lsa_implement_RNN}. Bridging the gap between Transformers' computational power and RNNs' inference efficiency is a thriving research area \citep{fournier_practical_2023}, and the link we made facilitates interpolation between those two model classes.

Finally, our work carries implications beyond deep learning. Inspired by evidence from neuroscience supporting the existence of synaptic plasticity at different timescales, previous work \citep{schmidhuber_learning_1992, ba_using_2016, miconi_differentiable_2018} added a fast Hebbian learning rule, akin to linear self-attention, to slow synaptic plasticity with RNNs. We show that, to some extent, this mechanism already exists within the neural dynamics, provided that the response of neurons can be multiplicatively amplified or shut-off in an input-dependent manner. Our results therefore suggest that recurrent neural circuits with long integration time constants, such as those found in the prefrontal cortex, might be learning and holding associations between past inputs in working memory. These circuits would effectively encode associative weights in their neural activity, not in actual synaptic connections, as would be the case for classical associative memory networks \citep{Steinbuch61,willshaw1969non,kohonen1972correlation}.  Interestingly, several single-neuron and circuit-level mechanisms have been experimentally identified which could support the required multiplication operation in biological neural networks \citep{silver_neuronal_2010}. We speculate that such multiplicative mechanisms could be involved in implementing self-attention-like computations in biological circuitry.

\subsection*{Acknowledgements}
The authors thank Asier Mujika and Razvan Pascanu for invaluable discussions. This study was supported by an Ambizione grant (PZ00P3\_186027) from the Swiss National Science Foundation and an ETH Research Grant (ETH-23 21-1).

%\subsection*{Impacts statement}

%This paper presents work whose goal is to advance the field of Machine Learning. There are many potential societal consequences of our work, none which we feel must be specifically highlighted here.

\newpage
\bibliography{refs}
\bibliographystyle{icml2024}

\clearpage
\appendix

\section{Additional details about the construction}
    
In Section~\ref{sec:construction} and Figure~\ref{fig:construction}, we have shortly described our construction. We here provide additional details, as well as refine it to settings in which we assume additional structure on the key, query and values matrices. We recall the mathematical definition of the gated RNN we consider:
\begin{align}
    h_{t+1} &= \lambda \odot h_t + g^\mathrm{in}(x_t)\\
    y_t &= D g^\mathrm{out}(h_t) \\
    g^\mathrm{in}(x) &= (W_\mathrm{m}^\mathrm{in} x) \odot (W_\mathrm{x}^\mathrm{in} x)\\
    g^\mathrm{out}(x) &= (W_\mathrm{m}^\mathrm{out} x) \odot (W_\mathrm{x}^\mathrm{out} x).
\end{align}

\subsection{Explicit values of the matrices of the vanilla construction}
\label{app:explicit-construction}

Here, we detail the values the matrices in Figure~\ref{fig:construction} take to mimic a linear self-attention layer with key, query and value matrices $W_K$, $W_Q$ and $W_V$. The key-values are stored in the first $d^2$ recurrent neurons and the queries in the last $d$ ones (indices $d^2+1$ to $d^2 + d$).

\paragraph{Input gating.} $W_\mathrm{x}^\mathrm{in}$ and $W^\mathrm{in}_\mathrm{m}$ are matrices of size $(d^2 + d) \times (d+1)$. The matrix $W_\mathrm{x}^\mathrm{in}$ both computes the values and the queries:
\begin{equation}
    {(W_\mathrm{x}^\mathrm{in})}_{i,j} = \left \{ 
    \begin{array}{ll} 
        {(W_V)}_{i/d,j} &\text{ if } j \leq d \text{ and } i \leq d^2 \\
        {(W_Q)}_{i-d^2,j} & \text{ if } j \leq d \text{ and } i > d^2 \\
        0 & \text{ otherwise}
    \end{array} 
    \right .
\end{equation}
and the matrix $W_\mathrm{m}^\mathrm{in}$ the keys:
\begin{equation}
    {(W_\mathrm{m}^\mathrm{in})}_{i,j} = \left \{ 
    \begin{array}{ll} {(W_K)}_{i\,\mathrm{mod}\,d,j} &\text{ if } j \leq d \text{ and } i \leq d^2 \\
    1 & \text{ if } j = d+1 \text{ and } i > d^2\\
    0 & \text{ otherwise}
    \end{array} 
    \right .
\end{equation}
where $/$ denotes integer division and $\mathrm{mod}$ the modulo operation.
As a consequence, the input received by the $i$-th recurrent neuron is $(W_V x)_{i / d} (W_K x)_{i~\mathrm{mod}~d}$ when $i \leq d^2$, and $(W_Q x)_{i - d^2}$ when $i > d^2$.
    
\paragraph{Recurrent neurons.} $\lambda$ is a vector of size $d^2 + d$ with
\begin{equation}
    \lambda_i = \left \{
    \begin{array}{ll}
    1 & \text{ if } i \leq d^2\\
    0 & \text{ otherwise.}
    \end{array}
    \right .
\end{equation}
The memory neurons, the first $d^2$ for which $\lambda = 1$, perfectly integrate all the key-values pairs.
    
\paragraph{Output gating.} $W^\mathrm{out}_\mathrm{x}$ and $W^\mathrm{out}_\mathrm{m}$ are matrices of size $d^2 \times (d^2 + d)$ with $W^\mathrm{out}_\mathrm{x}$ selecting the desired key-value element
\begin{equation}
    {(W_\mathrm{x}^\mathrm{out})}_{i,j} = \left \{ 
    \begin{array}{ll} 1 &\text{ if } j \leq d \text{ and } i = j \\
    
    0 & \text{ otherwise}
    \end{array} 
    \right .
\end{equation}
and $W^\mathrm{out}_\mathrm{m}$ the query element
\begin{equation}
    {(W_\mathrm{m}^\mathrm{out})}_{i,j} = \left \{ 
    \begin{array}{ll}
    1 & \text{ if } j > d^2 \text{ and } i = j \, \mathrm{mod} \, d \\
    0 & \text{ otherwise}
    \end{array} 
    \right .
\end{equation}
After the $d^2$ output neurons of the output gating thus contains all the $\left (\sum_{t'} (W_V x_{t'})(W_K x_{t'})^\top \right )_{i, j} (W_Q x_t)_{j}$ elements, and it only remains to sum them. 
    
\paragraph{Readout.} The goal of the readout matrix $D$, which has size $d \times d^2$, is to sum the key-values query products. It is equal to 
\begin{equation}
    D_{i,j} = \left \{
    \begin{array}{ll}
        1 & \text{ if } i = j / d\\
        0 & \text{ otherwise }
    \end{array}
    \right .
\end{equation}
The output $i$ of the gated RNN will thus be $\sum_j \left (\sum_{t'} (W_V x_{t'})(W_K x_{t'})^\top \right )_{i, j} (W_Q x_t)_{j}$, which is equals to $\left(\left (\sum_{t'} (W_V x_{t'})(W_K x_{t'})^\top \right ) (W_Q x_t) \right)_i$, the desired output.

\begin{figure}
    \centering
    \includegraphics{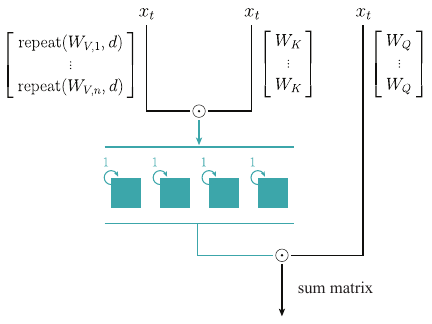}
    \caption{Construction for gated RNNs with side gating, as described in Section~\ref{app:side_gating_construction}}
    \label{fig:side_gating_construction}
\end{figure}

\subsection{Alternative construction with side gating}
\label{app:side_gating_construction}

With input and output gating, one has to waste some of the recurrent neurons to instantaneously pass through the query values. We chose this architecture because it is arguably simple and more common, but it is possible to give a RNN with a stronger inductive bias towards linear self-attention by replacing the output gating with a side gating, that is
\begin{equation}
    y_t = Dg^\mathrm{side}(x, h_t), ~~\mathrm{with}~~ g^\mathrm{side}(x, h) = (W^\mathrm{side} x) \odot h.
\end{equation}
Interestingly, this kind of side gating is featured in the recently proposed Mamba layer and indirectly in LSTMs, as we shall discuss in further detail in Section~\ref{app:comparison_architectures}. We detail how to adapt our construction to the side gating and provide a visual depiction of it in Figure~\ref{app:side_gating_construction}. Crucially, this construction only requires $\mathcal{O}(d^3)$ parameters instead of the $\mathcal{O}(d^4)$ of the previous one.

\paragraph{Input gating and recurrent neurons.} The construction remains the same as the previous one, except that we get rid of the constant term in the input and the last $d$ recurrent neurons.

\paragraph{Side gating. } The side gating matrix $W^\mathrm{side}$ is of size $\mathbb{R}^{d^2 \times d}$ has to copy queries $d$ times and put them in front of the corresponding key-value entry, that is
\begin{equation}
    W^\mathrm{side}_{i,j} = (W_Q)_{i~\mathrm{mod}~d, j}
\end{equation}

\paragraph{Readout matrix.} It remains the same as before.

\subsection{Reducing construction size with invertible $W_V$ / $W_K$}

In Section~\ref{subsec:number_neurons_construction}, we have argued that it is possible to reduce the number of recurrent neurons to $d(d+1)/2 + d$ when $W_Q$ is invertible. We use two insights.

\paragraph{Invariances of the linear self-attention layer. } The first thing we can remark is that modifying $W_Q$ and $W_K$ does not change the output of the layer as long as $W_K^\top W_Q$ is kept constant. This is because
\begin{multline}
    \left (\sum_{t'} (W_V x_{t'}) (W_K x_{t'})^\top \right ) (W_Q x_t)\\
    = W_V \left (\sum_{t'} x_{t'} x_{t'}^\top \right ) W_K^\top W_Q x_t
\end{multline}
It follows that a linear self-attention layer with weights $(W_K, W_Q, W_V)$ behaves similarly to one with weights $(W_V, W_V^{-\top} W_K^\top W_Q, W_V)$, as
\begin{equation}
W_V^\top W_V^{-\top} W_K^\top W_Q = W_K W_Q.
\end{equation}
Note that a similar argument holds if $W_K$ is invertible.

\paragraph{Symmetry of the key-values.} In the paragraph above, we have justified why we can consider the key and query values to be equal. In this case, the key-values matrix becomes symmetric. Knowing the elements contained in the upper triangular part is thus enough to know the entire matrix. We can thus ignore recurrent neurons corresponding to the lower triangular part. Note that similar insights apply to the side gating construction.

\subsection{Reducing construction size with low-rank teacher}
\label{app:low-rank-teacher}

Intuitively, when the teacher attention layer is of low rank, it is not necessary to represent all the elements of the key-values matrices if we can change the basis considered. We formalize this argument in the following. To that extent, we introduce the SVD decomposition of the value and query-key matrices:
\begin{align}
    W_V &= U_V \Sigma_V V_V^\top\\
    W_K^\top W_Q &= U_{KQ} \Sigma_{KQ} V_{KQ}^\top.
\end{align}
with $\Sigma$ diagonal matrices with as many non-zero elements as the rank of the matrix, and $U$ and $V$ orthogonal matrices.
The output of the attention layer can thus be written as
\begin{equation}
    U_V \left (\sum_{t'} (\Sigma_V V_V x_{t'}) (\Sigma_{KQ} U_{KQ} x_{t'})^\top \right ) V_{KQ} x_t.
\end{equation}
With this decomposition, only the first $\mathrm{rank}(W_V)$ rows and $\mathrm{rank}(W_K^\top W_Q)$ columns of the key-values matrix are not 0, that is we can reduce the number of recurrent neurons in our construction to $\mathrm{rank}(W_K^\top W_Q) \, \mathrm{rank}(W_V)$. Regarding the queries, only the first $\mathrm{rank}(W_K^\top W_Q)$ coordinates will be considered. In total, we thus need at most $\mathrm{rank}(W_K^\top W_Q) (\mathrm{rank}(W_V) + 1)$ neurons to replicate the teacher. As in the previous section, similar insights applies to the side gating construction.

To confirm that gated RNNs learn this solution, we performed a similar analysis to the one we did in Figure~\ref{fig:varying_student}.A, this time with low-rank teacher. To that extent, we take $d=12$ and restrict the rank of the key, query and value matrices to be 6. We do so by randomly sampling $W_K$, $W_Q$ and $W_V$ and removing $12 - 6 = 6$ singular values. Given the random sampling, $\mathrm{rank}(W_K^\top W_Q) = 6$ almost surely. We observe the stereotypical transition when the number of hidden neurons match $\mathrm{rank}(W_K^\top W_Q) (\mathrm{rank}(W_V) + 1) = 6 \times 7 = 42$, as plotted in Figure~\ref{fig:varying_student}.B.

\section{Gated RNNs and linear self-attention}
\label{app:comparison_architectures}

In this section, we compare our simplified gated RNN model, linear self-attention, and nonlinear gated RNN models (LSTMs, GRUs, LRUs and Mamba). We recall that the key ingredients of our simplified gated RNNs defined as
\begin{equation}
    h_{t+1} = \lambda \odot h_t + g^\mathrm{in} (x_t), ~~~ y_t = D g^\mathrm{out}(h_t),
\end{equation}
are the diagonal linear recurrence and the input and output gating. The input gating serves as a way to generate the key-values of linear self-attention, which will then be accumulated in the hidden recurrent units and combined with queries within the output gating.

Table \ref{tab:comparison_rnn_construction} summarizes how many layers of LRUs, Mamba, LSTMs and GRUs are needed to exactly implement our simplified class of gated RNNs and linear self-attention. We provide more details below.
\begin{table}[h]
    \centering
    \begin{tabular}{lp{1.75cm}p{1.75cm}}
        \toprule
         & Simplified gated RNN & Linear self-attention \\
         \midrule
         LRU & 2 & 2 \\ 
         LRU In-Out & 1 & 1 \\
         LRU In-Out (MLP) & -- & --\\
         Mamba & 2 & 1 \\
         LSTM & 2 & 1 \\
         GRU & -- & -- \\
         \bottomrule
    \end{tabular}
    \caption{Number of layers needed for different RNN layers to exactly implement our simplified class and linear self-attention.}
    \label{tab:comparison_rnn_construction}
\end{table}

\subsection{LRU}
\label{subsec:LRU}

An LRU layer \citep{orvieto_resurrecting_2023} consists of a recurrent state $h_t$ and some instantaneous post-processing. Its recurrent state is updated as
\begin{equation}
    h_{t+1} = \lambda \odot h_t + \gamma \odot (B x_{t+1})
\end{equation}
and its output $y_t$ is computed with
\begin{align}
    \tilde{y}_{t+1} &= \mathrm{Re}[Ch_t] + Dx_{t+1}\\
    y_{t+1} &= \sigma(W_\mathrm{m} \tilde{y}_{t+1}) \odot (W_\mathrm{x} \tilde{y}_{t+1}).
\end{align}
In the equations above, $h_{t+1}$, $B$ and $C$ are complex-valued, $\mathrm{Re}$ denotes the real part of a complex number, and $\sigma$ is the sigmoid function. The transformation nonlinear transformation between $y_{t+1}$ and $\Tilde{y}_{t+1}$ is called a gated linear unit (GLU) and was introduced in \cite{dauphin_language_2017}. Additionally, $\lambda$ and $\gamma$ are parametrized exponentially:
\begin{equation}
    \lambda = \exp(-\exp(\nu^\mathrm{log}) + i \exp(\theta^\mathrm{log})) ~~ \mathrm{and} ~~ \gamma = \exp(\gamma^\mathrm{log}).
\end{equation}

The LRU layer detailed above comprises two central computational mechanisms: a linear recurrence coupled with a GLU serving as nonlinear output gating. The recurrence is here complex-valued, but we only need the real part of it for our purposes. Assuming that the sigmoid can be linearized, our class of gated RNNs can be implemented using two layers by letting the output gating of the first layer serve as input gating. We are now left with linearizing the sigmoid. To achieve this, we double the number of output neurons of the GLU and require small weights in $W_\mathrm{m}$, that can for example, be compensated by large weights in $W_\mathrm{m}$. Under this regime, we have $\sigma(W_\mathrm{m} x) \odot (W_\mathrm{x} x) \approx (1/2 + W_\mathrm{m}x) \odot (W_\mathrm{x} x)$. Half of the neurons require identical weights as the target linear gating (up to a proportional factor), half should have $W_\mathrm{m} = 0$ and the same $W_\mathrm{x}$ as target linear gating. The $1/2 W_\mathrm{x} x$ term that comes from the second half of the neurons can be subtracted from the first half of the neurons in a subsequent linear transformation, thereby yielding the desired result.

In our experiments, we consider two additional variations of the LRU layer that can implement our class of gated RNNs and/or linear self-attention using only one layer. The LRU In+Out variation has an additional nonlinear input gating mechanism compared to the original version (LRU Out) that modifies the input before the recurrent part of the layer. The LRU In+Out (MLP) replaces the GLU in the LRU In-Out variation by a 1-hidden layer MLP, keeping the number of parameters fixed. The LRU In-Out variation can implement both linear self-attention and our class of gated RNNs in one layer, whereas LRU In-Out (MLP) cannot, as it does not have any multiplicative interactions.

\subsection{Mamba}

A (simplified) Mamba layer is defined as
\begin{align}
    \tilde{x}_{t} &= W^\mathrm{input}(x_t)\\
    \bar{A}_{t} &= \exp(\Delta(\tilde{x}_{t}) A(\tilde{x}_{t}))\\
    \bar{B}_{t} &= \Delta(\tilde{x}_{t}) B(\tilde{x}_{t})\\
    h_{t+1} &= \bar{A}_{t+1} h_t + \bar{B}_{t+1} \tilde{x}_{t+1}\\
    y_t &= C(\tilde{x}_{t+1})h_{t+1} \odot \sigma(W^\mathrm{side}(x_t))
\end{align}
where $\Delta$, $A$, $B$, $C$, $W^\mathrm{input}$ and $W^\mathrm{side}$ are linear transformations that produce resp. a scalar, matrix, matrix, matrix, vector and vector of appropriate size. For simplicity, we have ignored the convolutional layer after $W^\mathrm{input}$ in $\tilde{x}$, the fact that each coordinate of $\tilde{x}$ has its own independent recurrent layer and the specific parametrizations of the different parameters.

Here, the recurrence is linear with input-dependence, and thus more general than the one we are focusing on in this paper. It is easy to set it to what our construction requires. However, finding an input/output gating in this architecture is more tricky. The main insight is to look at
\begin{align}
    (B(x) x)_{i} &= \sum_j B(x)_{ij} x_j \\
    &= B(x)_{ii} x_i + \sum_{j \neq i} B(x)_{ij} x_j
\end{align}
and realize that it can implement a gating mechanism in which one of the branch is the identity. If it is preceded by a liner layer, such as $W^\mathrm{input}$ it can thus behave as the kind of gating we are focusing on in this paper. The input-dependent $B$ thus provides an input gating. The side gating we studied in Appendix~\ref{app:side_gating_construction} can be implemented through the side modulation, by linearizing the sigmoid, or indirectly through $C$. This implies that one single Mamba layer can emulate a linear self-attention layer. However, there is no mechanism to implement an output gating, so 2 layers are needed to mimick our simplified class of gated RNNs.

\subsection{LSTM}
An LSTM cell \citep{hochreiter_long_1997} has two recurrent states: the hidden state $h_t$ and the cell state $c_t$. They are updated as follows.
\begin{align}
    f_{t+1} &= \sigma(U_f x_{t+1} + V_f h_t + b_f)\\
    \Tilde{c}_{t+1} &= \tanh(U_c x_{t+1} + V_c h_t + b_c)\\
    g_{t+1} &= \sigma(U_g x_{t+1} + V_g h_t + b_g)  \label{eq:lstm_g}\\
    c_{t+1} &= f_{t+1} \odot c_t + g_{t+1} \odot \Tilde{c}_{t+1}  \label{eq:lstm_c}\\
    o_{t+1} &= \sigma(U_o x_{t+1} + V_o h_t + b_o)\\
    h_{t+1} &= o_{t+1} \odot \tanh(c_{t+1})  \label{eq:lstm_h}.
\end{align}
Here, $f_t$ is the cell state forget gate, $\Tilde{c}_t$ the cell state update candidate, $g_t$ the cell state update candidate gate, $o_t$ the output gate, and $\sigma$ the sigmoid function applied elementwise.

First, we show that one single LSTM layer can implement linear self-attention, by using $g_{t+1} \odot \Tilde{c}_{t+1}$ as a way to compute key-values and $c$ to aggregate them, $f_{t+1}$ and use $o_{t+1}$ for the query. We provide the corresponding weights in the table below, ignoring all the nonlinearities except $\sigma$ in the $f$ computation. Note that, compared to our simplified gated RNN class, we do not need to include neurons that forget their last state ($\lambda = 0$) here as the output gate directly provides the query to the output. Finally, linearizing the $\tanh$ function requires small $U_c$ weights that can later be compensated by large decoder weights, and ways to linearize the sigmoid were discussed in the previous section.

Implementing a gated RNN as in Equation~\ref{eq:gated_rnn} can be done by using two layers: in the first layer $g_{t+1} \odot \Tilde{c}_{t+1}$ serves as input gating, $f_{t+1}$ corresponds to $\lambda$, and, in the second layer, $g_{t+1} \odot \Tilde{c}_{t+1}$ serves as output gating. Table~\ref{tab:app_construction_LSTM} provides one set of such weights. This ignores the linearization trick for the $\tanh$ in $\tilde{c}$ and the sigmoid in $g_{t+1}$.

\begin{table*}[ht]
    \centering
    \begin{tabular}{cccc}
        \toprule
         & \multicolumn{3}{c}{Layer 1}\\
         \cmidrule(lr){2-4}
         & $U$ & $V$ & $b$ \\
        \midrule
        $f$ & $0$ & $0$ & $+\infty$ \\
        $\Tilde{c}$ & $\Tilde{W}_K$ & $0$ & $0$ \\
        $g$ & $\Tilde{W}_V$ & $0$ & $0$ \\
        $o$ & $\Tilde{W}_Q$ & $0$ & $0$\\
        \bottomrule
    \end{tabular}
    \hspace{0.5cm}
    \begin{tabular}{ccccccc}
        \toprule
         & \multicolumn{3}{c}{Layer 1} & \multicolumn{3}{c}{Layer 2} \\
         \cmidrule(lr){2-4}
         \cmidrule(lr){5-7}
         & $U$ & $V$ & $b$ & $U$ & $V$ & $b$ \\
        \midrule
        $f$ & $0$ & $0$ & $\sigma^{-1}(\lambda)$ & $0$ & $0$ & $-\infty$ \\
        $c$ & $W_\mathrm{m}^\mathrm{in}$ & $0$ & $0$ & $W_\mathrm{m}^\mathrm{out}$ & $0$ & $0$ \\
        $g$ & $W_\mathrm{x}^\mathrm{in}$ & $0$ & $0$ & $W_\mathrm{x}^\mathrm{out}$ & $0$ & $0$ \\
        $o$ & $0$ & $0$ & $+\infty$ & $0$ & $0$ & $+\infty$ \\
        \bottomrule
    \end{tabular}
    \caption{LSTM weight configuration that matches a linear self-attention layer (left) and a gated RNN as in Equation~\ref{eq:gated_rnn} (right). This presumes that the activation functions in $\tilde{c}$, $g$ and $o$ are linear. We use $\tilde{W}$ to denote the value, key and query matrices transformed in a similar way to what we did in Figure~\ref{fig:construction}.}
    \label{tab:app_construction_LSTM}
\end{table*}

\subsection{GRU}

A GRU cell \citep{cho_properties_2014} has a hidden state $h_t$, updated through
\begin{align}
    r_{t+1} &= \sigma(U_r x_{t+1} + V_r h_t+b_r)\label{eq:gru_r}\\
    \Tilde{h}_{t+1} &= \tanh(U_h x_{t+1} + V_h (r_{t+1} \odot h_t)+b_h)\label{eq:gru_ht}\\
    z_{t+1} &= \sigma(U_z x_{t+1} + V_z h_t+b_z)\label{eq:gru_z}\\
    h_{t+1} &= (1 - z_{t+1}) \odot h_t + z_{t+1} \odot \Tilde{h}_{t+1}\label{eq:gru_h}
\end{align}
where $r_t$ is the reset gate, $z_t$ is the update gate, $\Tilde{h}_t$ the update candidate, and $\sigma$ is the sigmoid function.

Here, stacking multiple GRUs on top of each other does not enable the implementation of any network from our class of gated RNNs nor linear self-attention layers. One layer can implement diagonal linear recurrence by linearizing the $\tanh$, having $z_{t+1} = 1$ and $r_{t+1} = \lambda$. However, implementing a gating mechanism of the form $g(x) = (W_\mathrm{m}x \odot W_\mathrm{x}x)$ is not possible\footnote{When the $\tanh$ is replaced by $\mathrm{Id}$, it is possible to achieve so by having $h_t \ll \Tilde{h}_{t+1}$ and correcting for the exponential growth in the next layer.}: we would need to use $z_{t+1}$ to implement one branch of the gating and $\tilde{h}_{t+1}$ the other but, given that $z_{t+1} \neq 0$, the previous hidden state $h_t$ influence the result.

\subsection{Can linear self-attention implement gated recurrent networks?}
\label{app:lsa_implement_RNN}

Throughout the paper, we mainly focus on understanding whether diagonal gated RNNs implement linear self-attention. In this section, we ask the opposite question: can linear self-attention layers can implement gated recurrent networks. The answer is that attention layers as we defined in Section~\ref{subsec:background_lsa} cannot, because it can only perfectly integrate inputs or send the current one (thus $\lambda = 0$ or $\lambda = 1$). However, adding a mechanism akin to weight decay bridges the gap. In particular, we will describe how the output $y_t$ of a such a linear self-attention layer can satisfy a recurrence relationship of the form $y_{t+1} = \lambda \odot y_t + x_t$. To do so, we consider the following attention layer:
\begin{align}
    v_t &= W_V x_{t} + b_V\\
    k_t &= W_K x_{t} + b_K\\
    q_t &= W_Q x_t + b_Q\\
    y_t &= \left ( \sum_{t'=1}^t \Gamma_{t-t'} \odot (v_{t'} k_{t'}^\top) \right ) q_t
\end{align}
where $\Gamma_{t-t'}$ is a matrix of size $d\times d$ in which all entries of the $i$-th row have value $(1 - \gamma_i)^{t-t'}$. Such a layer is featured in recent work, e.g. \cite{sun_retentive_2023} or \cite{yang_gated_2023}. The $\gamma$ term can be interpreted as a weight decay: if we note 
\begin{equation}
    W^\mathrm{ff}_t := \left ( \sum_{t'=1}^t \Gamma_{t'-t} \odot (W_V x_{t'}) (W_K x_{t'})^\top \right ) \! ,
\end{equation}
we have
\begin{equation}
    W^\mathrm{ff}_{t+1} = W^\mathrm{ff}_t + (W_V x_{t+1} + b_V) (W_K x_{t+1} + b_K)^\top - \Gamma_1 W^\mathrm{ff}_t.
\end{equation}
Now, we set the value, key and query matrices and biases to $W_V = \mathrm{Id}, b_V = 0, W_K = 0, b_K = 1, W_Q = 0, b_Q = 1/d$ and $1-\gamma = \lambda$. This way, we have
\begin{align}
    y_{t+1} &= \frac{1}{d} W^\mathrm{ff}_{t+1} 1\\
    &=  \frac{1}{d} \left ( \Gamma_1 \odot W^\mathrm{ff}_t + x_{t+1} 1^\top \right ) 1\\
    &= \left ( \Gamma_1 \odot W^\mathrm{ff}_t \right ) 1 + x_{t+1}\\
    &= \lambda \odot y_t + x_{t+1}
\end{align}
In the last line, we use the structure of $\Gamma_1$ and the value of $\gamma$. Biases terms are crucial to make this link: without them $W_t^\mathrm{ff}$ would be a polynomial with only degree 2 coefficients and the equivalence would not be possible. The gating mechanism within networks described in Equation~\ref{eq:gated_rnn} can also be implemented by forgetting ($1 - \gamma = 0$) and having the key-value taking care of the multiplication.

This analysis reveals the importance of weight decay to implement recurrent neural network like computations with a wide range of timescales. Adding complex-valued weight decay to linear self-attention layers makes them closer to state-of-the-art recurrent neural networks architecture \citep{orvieto_resurrecting_2023, smith_simplified_2023} for capturing long-range dependencies. Therefore, such a modification might boost the performance of attention layers on benchmarks testing these properties, such as the Long Range Arena \citep{tay_long_2020}. Interestingly, this view can partly explain the great empirical performance of the RWKV \citep{peng_rwkv_2023}, which features a similar mechanism to weight decay. Overall, the analysis we conducted in this section examplify how the connection between RNNs and attention layers we made in this paper can be used to guide development of future architectures.

\section{Teacher-student}
\label{app:teacher_student}

\subsection{Experimental details}
\label{app:TS-experimental-details}

For all experiments in Section~\ref{sec:gatedRNN_identification}, we train the student for almost one million training iterations on sequences of length 32 and a batch size of 64 (50000 training examples per epoch, 1000 epochs). We use the AdamW \citep{loshchilov_decoupled_2019} optimizer with a cosine annealing learning rate scheduler. The initial learning rate is set at $10^{-3}$, scheduled to anneal down to $10^{-6}$ by the end of training and a weight decay of $10^{-4}$ is applied to all parameters except the recurrent ones $\lambda$ in the experiment of Section~\ref{subsec:teacher-identification}. To ensure that the hidden states do not explode, we ensure that $\lambda$ stays within $[0, 1]$ by employing the exponential parametrization described in Appendix~\ref{subsec:LRU} (we only keep the $\nu$ part as $\lambda$ takes real values here).

\begin{figure*}
    \centering
    \includegraphics{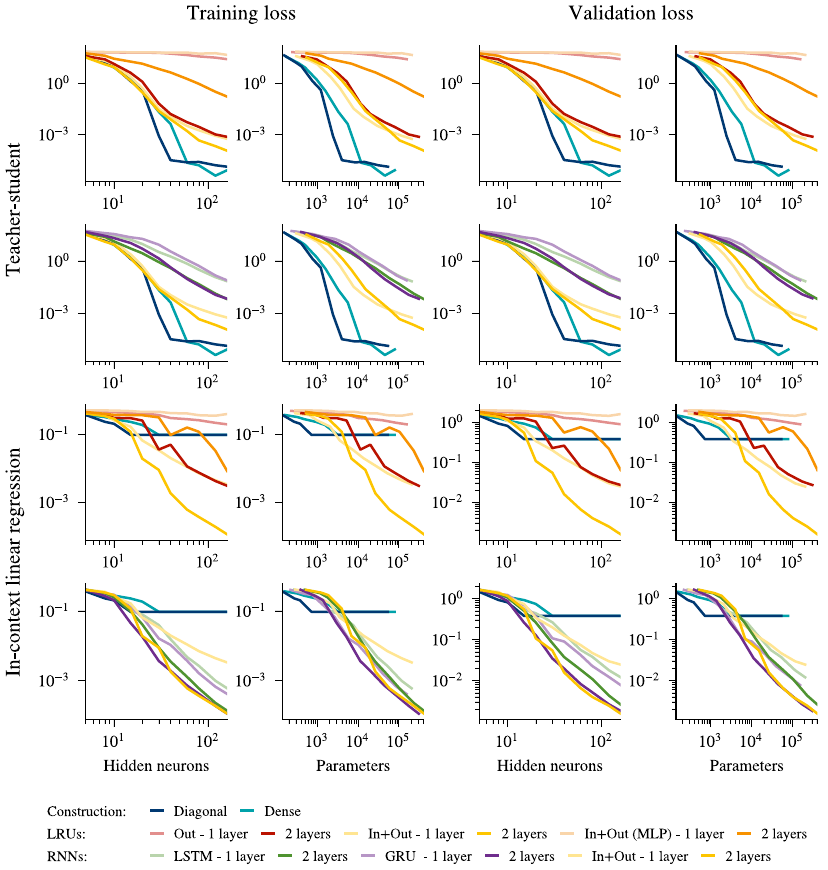}
    \caption{Extensive comparison between the different architectures. Compared to Figure~\ref{fig:comparison}, we consider different versions of the LRU here, plot the loss as the function of the number of parameters, and include both training and validation losses. Those two losses are almost (up to some sampling noise) for the teacher-student task but are different for the in-context linear regression task because we change the $W^*$ distribution in the validation set.}
    \label{fig:extensive_comparison}
\end{figure*}

In Figure~\ref{fig:extensive_comparison}, we add more results to the architecture comparison we did in Figure~\ref{fig:comparison}. In particular, we compare the three different types of LRU we mentioned in Appendix~\ref{subsec:LRU}, and observe that adding an input GLU improves LRUs ability to mimic linear self-attention within one layer, but also with several layers.

\subsection{Compression of the learned output gating weights}
\label{app:compression_output_gating}

In Figure~\ref{fig:result-learning}, we show that the gating weight matrices have a structure that is close to the one of our construction, except for three different rows (11, 12, and 13). We claim they can be reduced to a single row; we now provide details justifying it.

Therefore, our objective is to demonstrate that these three rows are functionally equivalent to a single row with the expected structure and to gain insights into the invariances inherent to the gating mechanism we study in this paper along the way. The initial step toward achieving this entails examining the influence of these three rows on the $i$-th coordinate of the network's output:
\begin{align}
    \sum_{j=11}^{13} D_{i,j} g^\mathrm{out}(h)_j &= \sum_{j=11}^{13} D_{i,j} (W^\mathrm{out}_{\mathrm{m},j} x)(W^\mathrm{out}_{\mathrm{x}, j} x)\\
    &= x^\top \left ( \sum_{j=11}^{13} D_{i,j} W^\mathrm{out}_{\mathrm{m},j} \,{W^\mathrm{out}_{\mathrm{x}, j}}^\top \right ) x.
\end{align}
This contribution can be interpreted as a quadratic form whose kernel is a weighted sum of rank-1 kernels defined by the rows of the output gating matrices. In Figure~\ref{fig:result-learning}.C, we plot the obtained kernel for one of the output components. Crucially, the resulting kernel for the four output units are all proportional to one another and is of rank-1. We can thus reduce the three neurons (11, 12 and 13) to one. Furthermore, the two vectors whose outer product yields the resulting kernel now mirror the construction's structure. One of these two vectors exclusively accesses query neurons while the other reads key-value neurons, as seen in Figure~\ref{fig:result-learning}.C.
As usually occurs with this kind of manipulation \citep{martinelli_expand-and-cluster_2023}, merging the neurons slightly increases the loss, but original loss levels can be recovered after fine-tuning.

\section{In-context linear regression} 

\subsection{Experimental details}
\label{app:ICL-experimental-details}

In the in-context linear regression experiment, each sequence is a task characterized by a unique $W^*$. The weight matrix $W^*$ entries are sampled i.i.d. from a normal distribution $\mathcal{N}(0,\frac{1}{3})$. Each element of the sequence is of the form $(x_t, W^* x_t)$. The entries of the inputs $(x_t)_{t=1}^{T+1}$ are sampled i.i.d. from the uniform distribution $\mathcal{U}(-\sqrt{3},\sqrt{3})$. During the validation phase, we draw tasks from a different distribution, $W^*_{ij} \sim \mathcal{N}(0,\frac{2}{3})$ to highlight the generalization abilities of the learned models. We train the model with the same optimization scheme described in Appendix~ \ref{app:TS-experimental-details}, except that we use a smaller number of training iterations, totaling $300,000$. By default, we use gated RNNs with 80 hidden neurons.

\subsection{Optimal learning rate for one-step gradient descent}
\label{sec:optimal_lr}

Let $X \in \R^{d_x \times n}, W \in \R^{d_y \times d_x}$ random variables such that all entries of $X$ are sampled i.i.d. from a centered uniform distribution with variance $\sigma_x^2$, and those of $W$ i.i.d. from some centered distribution with finite variance $\sigma_W^2$. We set $Y=WX$. Let $x \in \R^{d_y}$ a column vector, whose entries are sampled from the same distribution as those of $X$, and $y=Wx$.

The goal of this section is to analytically derive the optimal learning rate for the in-context linear regression task, that is to find $\eta$ which minimizes
\begin{equation}
    \mathcal{L}(\eta) = \frac{1}{2}\E_{X,W,Y,x,y} \left [ \| y - \hat{W}(\eta , X,Y)x \|^2 \right ]
\end{equation}
where $\hat{W}(X,Y)$ is the result of one gradient descent step starting from $0$ with learning rate $\eta$ on the loss $W \mapsto \frac{1}{2}\| Y - WX \|^2$. The calculation is presented in a more general form in \cite{mahankali_one_2023}. We include it here as we additionally provide a simple formula for  exact optimal learning rate value.

Plugging in the analytical expressions for $y$ and $\hat{W}$, we get
\begin{align}
    \mathcal{L}(\eta) &= \frac{1}{2}\E_{X,W,Y,x,y} \left [ \| y - \eta YX^\top x \|^2 \right ]\\
    &= \frac{1}{2}\E_{X,W,x} \left [ \| Wx - \eta WXX^\top x \|^2 \right ]\\
    &= \frac{1}{2}\E_{X,W,x} \left [ \| W(I - \eta XX^\top) x \|^2 \right ]
\end{align}

We want to minimize $\mathcal{L}$, i.e. look for $\eta^*$ that satisfies $\partial_\eta \mathcal{L}(\eta^*) = 0$. We have
\begin{align}
    \partial_\eta \mathcal{L}(\eta) &= \E_{X,W,x} \left [ \left (W(I - \eta XX^\top)x\right )^\top WXX^\top x \right ]\\
    &= \Tr\E_{X,W,x} \left [ (I - \eta XX^\top)W^\top W XX^\top x x^\top \right ]\\
    &= \sigma_x^2 \Tr\E_{X,W} \left [ (I - \eta XX^\top)W^\top W XX^\top \right ]\\
    &= \sigma_x^2 \Tr\E_{X,W} \left [XX^\top(I - \eta XX^\top)W^\top W \right ]\\
    &= \sigma_x^2 \sigma_W^2 \Tr\E_{X} \left [XX^\top(I - \eta XX^\top) \right ]
\end{align}
In the first equation, we use that $\E[a^\top b] = \Tr \E [b a^\top]$. Third and fifth ones make use of $\E_{x}[xx^\top] = \sigma_x^2 \mathrm{Id}$ and $\E_{W}[WW^\top ]=\sigma_W^2 \mathrm{Id}$. Having $\partial_\eta \mathcal{L}(\eta^*) = 0$ is then equivalent to
\begin{equation}
    \eta^\star \coloneqq \frac{\Tr\E_{X}[XX^\top]}{\Tr\E_{X}[XX^\top X X^\top]}.
\end{equation}

This result shows that only the distribution of the learning data matters. Let us compute this quantity. We have $\E_{X}[XX^\top ]=n\sigma_x^2 \mathrm{Id}$ so we are left with computing $\E_{x}[XX^\top X X^\top]$. Using that entries of $X$ are i.i.d., we get
\begin{align}
    &\Tr\E_{X}[XX^\top X X^\top] \\
    &= d_x \E_X \left [ \sum_i \left (\sum_t x_{i, t}x_{1, t}\right)^2\right]\\
    &= d_x \E_X\left[\left(\sum_t x_{1, t}^2\right)^2\right] \\
    & ~~~~~~~~~~ + d_x (d_x-1) \E_X\left[\left(\sum_t x_{1, t}x_{2, t}\right)^2\right] \\
    &= d_x \E_X\left[\sum_t x_{1, t}^4 +\sum_{t\neq t'} x_{1, t}^2x_{1, t'}^2\right]\\
    & ~~~~~~~~~~ + d_x(d_x-1)\E_X\left[\sum_t x_{2, t}^2x_{1, t}^2\right]\\
    &= \frac{9}{5} n d_x \sigma_x^4 + n(n-1)d_x \sigma_x^4 + n(d_x - 1)\sigma_x^4 \\
    &= n d_x \sigma_x^4 \left(n + d_x - \frac{1}{5}\right)
\end{align}
because the fourth moment of a centered uniform distribution is $\frac{9}{5}\sigma_x^4$. Putting everything together, we finally have
\begin{equation}
    \eta^* =\frac{1}{\sigma_x^2(n + d_x - \frac{1}{5})}.
\end{equation}

\subsection{Associative recall}
\label{app:assoc-recall}

As a complement to in-context linear regression, we consider a simple in-context classification task studied by \citet{fu_hungry_2023}, where the network has to remember associations between paired inputs. As for in-context regression, the network is presented with a sequence of tokens of the form $(x_t, y_t)_{1\leq t \leq T}$, followed by a token containing a query input and a null placeholder $(x_{T+1},0)$. In this task, $x_{T+1}$ corresponds exactly to one of the previously seen $x_t$, and the goal is to complete the placeholder with the corresponding $y_t$.

To make the task solvable by a single layer of linear attention, we present the following sequence: $([x_1, y_1], [y_1, x_2],  [x_2, y_2] \dots, [x_T, y_T], [x_{T+1}, 0])$, where $x$ (resp $y$) have been transformed to a $2T$-sized one-hot encoding of $[1,T]$ (resp. $[T+1, 2T]$), resulting in a input dimension of $2T$. Each $x$ and each $y$ only appear once. We use a cross entropy loss, using the desired $y$ as target, and $T=8$ in our experiments.

\textbf{Solving the task with linear self-attention. } Given that we provide non-repeating one hot encoded inputs, we can see that a linear self-attention layer that uses $x$ as key and query, and $y$ as value will solve the task. That is, its output is
\begin{equation}
    y_{T+1} = \left ( \sum_{t \leq T} y_t x_t^\top \right ) x_{T+1}.
\end{equation}

\textbf{Input-output gating.} We first trained a gated RNN on this task and observe that the solution it finds differs from the linear self-attention layer, and requires way less recurrent neurons. To each $y$, it associates a recurrent neuron with $\lambda=1$ in which it will store a value $v_x$ corresponding to $x$ when that $y$ appears. That is, if the pair $(x, y)$ appears, the recurrent neuron associated to $y$ receives $v_x$ as input, and the other receive no input. Addtionally, the RNN uses one neuron with $\lambda = 0$ containing the value $v_x$ associated to the current $x$. The output gating then computes the negative squared difference between the current value and the stored ones, so that neural activity after gating is equal to $(-(v_{x_{T+1}} - v_{x(y)})^2)_y$ where $x(y)$ is the $x$ that was associated to $y$ in the sequence. The index of the smallest one, equal to 0, gives the desired output after taking the argmax. We note that such a solution is possible as each $x$ and $y$ appear only once in the sequence, as this is a classification class and as inputs are one-hot encoded.

\textbf{Side gating.} Then, we use the RNN with side gating of Section~\ref{app:side_gating_construction}, with parameters $W^\mathrm{in}_\mathrm{x}$, $W^\mathrm{in}_\mathrm{m}$, $\lambda$ and $W^\mathrm{side}$, and check whether it implements the same function as the RNN with input-output gating. It does not, and we detail the solution it finds in the following. We apply the same post processing of the weights as we did in Section~\ref{sec:gatedRNN_identification}, and find that only recurrent neurons with $\lambda = 1$ are remaining. Consistently with the linear self-attention layer that optimally solves this task, one of the input gating matrix, $W^\mathrm{in}_\mathrm{x}$ on reads out from the $x$ part of the input, and the other one, $W^\mathrm{in}_\mathrm{m}$ from $y$. Additionally, the side gating matrix is equal to the $W^\mathrm{in}_\mathrm{x}$ matrix, in a similar way that the query matrix is equal the key one in the linear self-attention layer. Finally, the $D$ matrix is the transpose of the value-like part of matrix $W_\mathrm{m}^\mathrm{in}$. Based on those observations, we can rewrite
\begin{align}
    W_\mathrm{x}^\mathrm{in} = W^\mathrm{side} &= [A \, | \, 0]\\
    W_\mathrm{m}^\mathrm{in} &= [0 \, | \, B]\\
    D &= B^\top
\end{align}
As $\lambda = 1$, we have
\begin{equation}
    h_{T} = \sum_{t \leq T} g^\mathrm{in}([x_t, y_t]) = \sum_{t \leq T} (B y_t) \odot (A x_t)
\end{equation}
and
\begin{align}
    y_{T+1} &= h_{T+1} \odot (W x_{T+1})\\
    &= B^\top \sum_{t\leq T} (B y_t) \odot (A x_t) \odot (A x_{T+1})\\
    &= \sum_{t\leq T} M(x_t, y_t) x_{T+1}
\end{align}
In the last equation, we remarked that $y_{T+1}$ is a linear function of $x_{T+1}$ so that we can write it as a matrix, and this matrix a sum of matrices that depend linearly on $x_t$ and $y_t$.

\begin{figure*}
    \centering
    \includegraphics[width=0.9\textwidth]{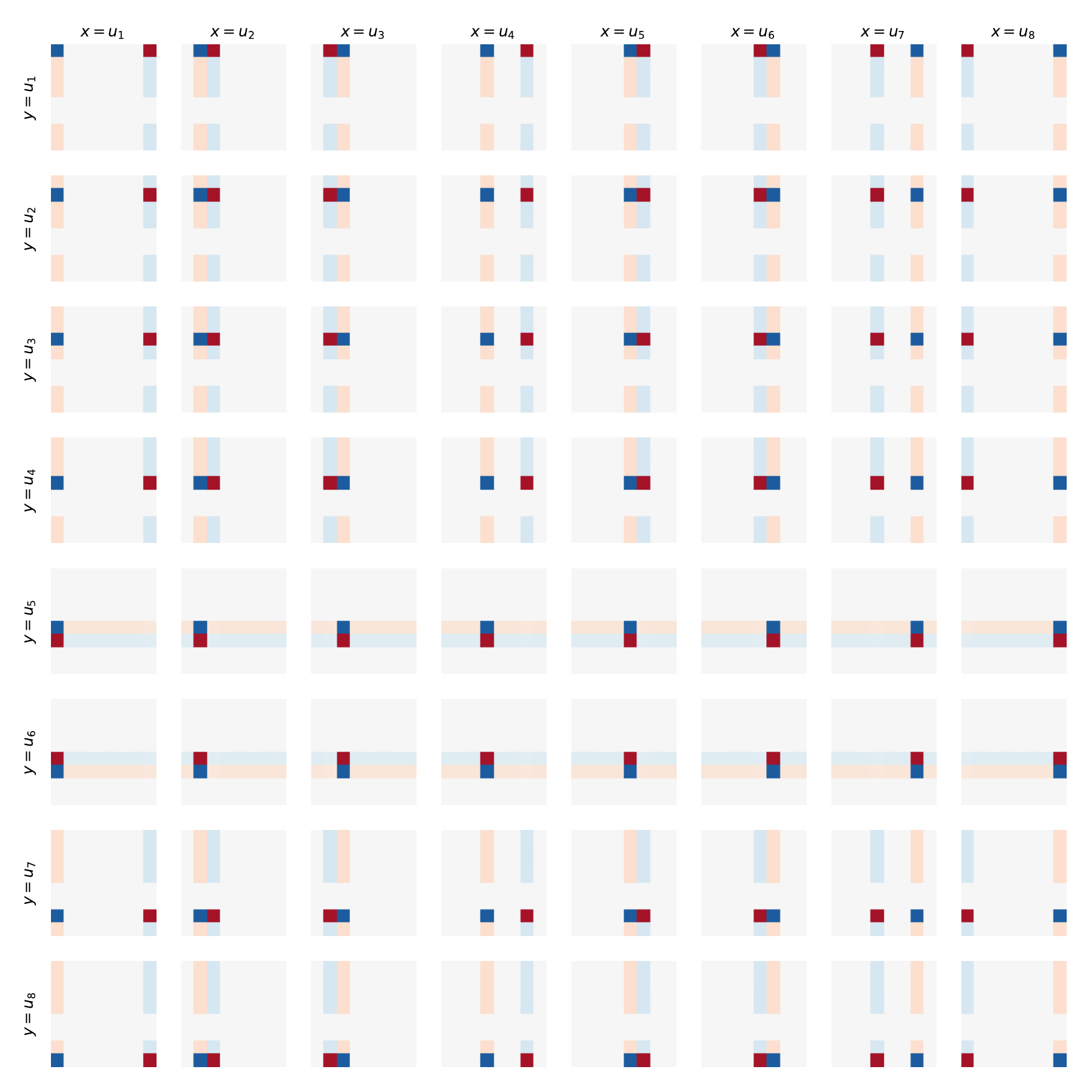}
    \caption{Values taken by the $M(x, y)$ when $x$ and $y$ are equal to the canonical basis. The obtained matrices are all of rank 1.}
    \label{fig:app_M_matrix}
\end{figure*}

We can now compare the behavior of this solution, with the solution found by linear self-attention, by looking in more detail into the $M$ matrices. We first observe that $M$ and $(x, y) \mapsto yx^\top$ are bilinear so that it is enough to study their behavior on the canonical basis $(u_i)_i$. We plot those different matrices on Figure~\ref{fig:app_M_matrix}. We observe that each component is of rank 1 similarly to the self-attention layer solution, with a peak on the component $(i, j)$ as expected. However, there is an additional negative peak, of same amplitude as the positive one, that does not affect the prediction as we are dealing with one-hot encoded inputs and outputs and a classification task. One putative reason to explain explain these observations is that, as the patterns are one-hot encoded, it is possible to represent them in $\log T$ neurons, without affecting classification performance. This would require less neurons in the output gating as what the link with attention would, and can be compensated with the kind of binary patterns we observe. Alternatively, binary patterns do not cover all directions from the input space so identification might become more difficult.

\section{Software}

We run our experiments using the Jax \citep{bradbury_jax_2018} Python framework, using the Flax \citep{heek_flax_2023} library for neural networks. We base our code base on the Minimal-LRU \citep{zucchet_minimal_2023} repository. Data analysis and visualization were done using Numpy \citep{harris_array_2020}, Scikit-learn \citep{pedregosa_scikit-learn_2011} and Matplotlib \citep{hunter_matplotlib_2007}.

\end{document}